\documentclass{article}

\usepackage[nonatbib,preprint]{neurips_2021}
\usepackage[square,numbers]{natbib}

\usepackage[utf8]{inputenc} 
\usepackage[T1]{fontenc}    
\usepackage{url}            
\usepackage{booktabs}       
\usepackage{amsfonts,amsmath,amssymb,amsthm, mathtools}       
\usepackage{nicefrac}       
\usepackage{microtype}      
\usepackage{xcolor}
\usepackage{scalerel}
\usepackage[noend]{algorithmic} 
\usepackage[algoruled]{algorithm2e}
\usepackage{censor}
\usepackage[most]{tcolorbox}

\DeclareMathOperator{\proj}{proj}

\DeclareMathSymbol{\shortminus}{\mathbin}{AMSa}{"39}

\newcommand{\Nrm}{\mathcal{N}}

\newcommand{\EE}{\mathbb{E}}
\newcommand{\II}{\mathbb{I}}
\newcommand{\inv}{^{-1}}

\newcommand{\data}{\mathcal{D}}

\newcommand{\util}{\mathfrak{u}}
\newcommand{\Util}{U}
\newcommand{\EEutil}{\mathcal{U}}

\newcommand{\ksx}{\mathbf{k}_*}
\newcommand{\kxs}{\mathbf{k}_*^\top}
\newcommand{\kss}{k_{**}}
\newcommand{\Kxx}{K}

\newcommand{\ys}{y_*}
\newcommand{\fs}{f_*}
\newcommand{\bx}{\mathbf{x}}
\newcommand{\bxs}{\mathbf{x}_*}
\newcommand{\by}{\mathbf{y}}

\newcommand{\bff}{\mathbf{f}}
\newcommand{\bmu}{\boldsymbol{\mu}}

\newcommand{\nat}{{\theta_1}}
\newcommand{\Nat}{{\theta_2}}
\newcommand{\mean}{{\eta_1}}
\newcommand{\Mean}{{\eta_2}}
\newcommand{\opt}{\text{opt}}

\newcommand{\cav}[1]{_{\backslash #1}}
\newcommand{\scav}[1]{^{\backslash #1}}
\newcommand{\new}{^{\text{new}}}

\newcommand{\param}{\varphi}
\newcommand{\Tcrit}{\tau_\text{crit}}

\definecolor{sidenote-gray}{HTML}{494649}
\definecolor{less-pale-gray}{gray}{0.9}
\definecolor{pale-gray}{gray}{0.98}
\definecolor{mydarkblue}{rgb}{0.1,0.1,0.7} 
\usepackage[colorlinks=true,allcolors=mydarkblue]{hyperref}

\title{Loss-calibrated expectation propagation for approximate Bayesian decision-making}

\author{%
  Michael J.~Morais \& Jonathan W.~Pillow\\
  Princeton Neuroscience Institute\\
  Princeton University\\
  \texttt{mjmorais,~pillow@princeton.edu} \\
}

\begin{document}

\maketitle

\begin{abstract}
Approximate Bayesian inference methods provide a powerful suite of tools for finding approximations to intractable posterior distributions. However, machine learning applications typically involve selecting actions, which---in a Bayesian setting---depend on the posterior distribution only via its contribution to expected utility. A growing body of work on ``loss-calibrated approximate inference'' methods has therefore sought to develop posterior approximations sensitive to the influence of the utility function. Here we introduce loss-calibrated expectation propagation (Loss-EP), a  loss-calibrated variant of expectation propagation. This method resembles standard EP with an additional factor that ``tilts'' the posterior towards higher-utility decisions. We show applications to Gaussian process classification under binary utility functions with asymmetric penalties on False Negative and False Positive errors, and show how this asymmetry can have dramatic consequences on what information is ``useful'' to capture in an approximation.
\end{abstract}

\section{Introduction}

Bayesian methods tend to emphasize { inference}, which involves modeling the uncertainty over model parameters given an observed dataset. However, in most cases we perform Bayesian inference in order to make decisions or take actions in some end-use task \citep{von2012clustering}. 
Bayesian decision theory asserts that the posterior distribution learned from inference is sufficient for Bayes-optimal decision-making \citep{berger2013statistical}.  This theoretical result is of limited value if the posterior distribution is intractable---an increasingly common occurrence in modern machine learning settings with
large datasets and/or complex models.

To address this intractability, approximate Bayesian inference methods seek to compute tractable approximations to the posterior distribution.  It is tempting to use these approximations as a plug-in replacement for the posterior during prediction and decision-making. However, information is lost in these approximations, and our goal {\it should} be to retain information that allows us to make optimal decisions, i.e., decisions that maximize utility. Unfortunately, common approximate inference methods such as variational inference (VI) and expectation propagation (EP) do not take account of utility, or of the loss of information that could compromise optimal decision-making.
The key intuition is that if we change the utility function, we do not change the posterior distribution, but we may change the best approximation to it. Loss-calibrated approximate inference has only recently emerged as a general framework optimizing and evaluating approximations in a ``loss-aware'' way \citep{lacoste2011approximate}. Analogously to how Bayesian decision theory prescribes that optimal decision-making under uncertainty maximizes expected utility---an expectation over the posterior distribution of the models' parameters---loss-calibrated inference proposes that the optimal approximation too should maximize expected utility. 


With this paper, we introduce loss-calibrated expectation propagation (loss-EP). Like other loss-calibrated inference methods, loss-EP aims to embed features of the decision-making problem into the inference procedure. We show  loss-EP is tantamount to standard EP  augmented by one additional site factor that ``calibrates'' the approximate posterior to the decision-making problem. In Section 2, we outline the core machinery of Bayesian decision theory and loss-calibrated inference. In Section 3, we introduce EP and loss-calibrated EP. In Section 4, we demonstrate the performance of loss-calibrated and standard approximate inference methods in the familiar Gaussian process classification. In Section 5, we discuss opportunities for future work that could complement and/or improve Loss-EP as presented here.

\section{Background}
\subsection{Bayesian decision theory}
The Bayesian approach to decision-making starts from a model with parameters $\param$, likelihood $p(y\mid \bx, \param)$, and prior $p(\param)$. Given an observed training dataset $\data=\{\bx_n, y_n\}_{n=1}^N$, the posterior follows from Bayes' rule:
\begin{flalign}
\label{eq:generic-posterior}
p(\param\mid\data)=\frac{p(\param)p(\data\mid\param)}{p(\data)}\propto p(\param)\prod_{n=1}^N p(y_n\mid\bx_n,\param)
\end{flalign}
Given novel datapoints $\bxs$, we can make predictions (incorporating uncertainty) about future observations using the posterior predictive distribution, which incorporates all relevant information from the training data via the posterior: 
\begin{flalign}
\label{eq:generic-prediction}
p(\ys \mid \bxs, \data)=\int p(\ys \mid \bxs, \param) p(\param \mid \data) d\param
\end{flalign}
The end-use decision-making problem selects an action $a(\bxs)$ and receives a utility $\util(\ys, a(\cdot))$, depending on the observation $\ys$ thereafter. To select the optimal action, we briefly suppose we can compute the posterior in eq. \eqref{eq:generic-posterior}. Bayesian decision theory prescribes that we choose the action(s) that maximize the full expected utility, inversely minimize the full Bayes risk. The full expected utility computes the utility received for future (predictive) data $(\bxs, \ys)$, in expectation over those posterior beliefs about the model parameters:
\begin{flalign}
\label{eq:full-expected-util}
\EEutil_p(a)&=\int p(\param\mid\data)\iint p(\bxs)p(\ys\mid\bxs,\param)\,\util(a(\bxs), \ys)\: d\ys d\bxs\cdot d\param \triangleq \int p(\param\mid\data) \Util(a, \param) d\param
\end{flalign}
where $\Util(a, \param)$ is the predictive utility, marginalized over data given a fixed value of model parameters. In practice, the data-generating distribution $p(\bxs)$ is often unknown, for which we can define the conditional expected utility given some $\bxs$ to be 
\begin{flalign}
\label{eq:cond-expected-util}
\EEutil_p(a(\bxs)\mid\bxs) = \iint p(\param\mid\data)p(\ys\mid\bxs,\param)\util(a(\bxs), \ys)\: d\ys\cdot d\param
\end{flalign}
The optimal action function $a_\opt$ maximizes the full expected utility in \eqref{eq:full-expected-util} as 
\begin{flalign}
a_{\text{opt}} = \underset{a}{\arg\min}\:\EEutil_p(a) \label{eq:opt-action}
\end{flalign} 
This would be a cumbersome optimization over several intractable integrals, but importantly it is equivalent to the optimal actions achieved by maximizing the conditional expected utility in (\ref{eq:cond-expected-util}) pointwise, such that $a_{\opt}(\bxs) = {\arg\min_a}\:\:\EEutil_p(a(\bxs)\mid\bxs)$ also.


\subsection{Loss-calibration as approximate action selection}

In most applications of interest, exact inference of the true posterior is intractable, which poses a major challenge for Bayes-optimal decision-making. Conceivably, the next-best thing we can do is to make actions optimal under some tractable approximation $q(\param)$ of the posterior $p(\param\mid\data)$. In such case, we could define the analogous approximate full and conditional expected utilities under the approximation; the latter, along with the action that maximizes it, would be
\begin{flalign}
\label{eq:q-cond-expected-util}
\EEutil_q(a(\bxs)\mid\bxs) &= \iint q(\param)p(\ys\mid\bxs,\param)\util(a(\bxs), \ys)\: d\ys\cdot d\param\\
\label{eq:q-action}
\text{with}\quad a_{q}(\bxs) &= \underset{a}{\arg\max}\:\:\EEutil_q(a(\bxs)\mid\bxs).
\end{flalign}
Whether or not this $q$-optimal action $a_q$ maximizes the true expected utility in \eqref{eq:full-expected-util} will depend on how we select the approximation $q(\param)$. Our approximate utility maximization using $q(\param)$ and $a_q$ in eqs. (\ref{eq:q-cond-expected-util}, \ref{eq:q-action}) will be suboptimal---relative to the Bayes-optimal $p(\param\mid\data)$ and $a_\text{opt}$---to the extent that for any $\bxs$, we select a suboptimal action, i.e. $a_q(\bxs)\neq a_\text{opt}(\bxs)$. 
In expectation, we can define the discrepancy in utility between $q$-optimal actions and true optimal actions as
\begin{flalign}
d_\util(p\,\|\, q) = \EEutil_p(a_\text{opt}) - \EEutil_p(a_q) \label{eq:regret}
\end{flalign}
This discrepancy is neither a proper divergence nor regret---there can be many $q$ and $a_q$ such that $d_\util=0$, but still serves as a useful guide for how we should define the ``best'' approximate posterior. Problematically, optimizing expected utility as a function of $q$ is not the objective of approximate inference methods. Loss-calibrated approximate inference was proposed to bridge this gap, and jointly perform approximate Bayesian inference and action selection, by incorporating the expected utility explicitly into the optimization program that performs inference.

\subsection{Variational inference and loss-calibrated variational inference}

For example, variational inference (VI) optimizes a bound on the marginal likelihood \citep{blei2017variational}:
\begin{flalign}
\log p(\data) &\ge \int q(\param)\cdot \log\left(\frac{p(\param,\data)}{q(\param)}\right)\,d\param \\
\text{s.t.}\quad q_\text{VI}(\param)&=\underset{q}{\arg\min}\: D_{KL}\left(p(\param\mid\data)\,\|\, q(\param)\right)
\end{flalign}
Minimizing KL divergence optimizes generic statistical features of the posterior, but such an approximation could still result in arbitrarily poor decisions or arbitrarily large discrepancy $d_\util(p\,\|\, q)$. Matching probability density is not important---matching actions is. 


Loss-calibrated variational inference (LossVI) approximates the posterior by maximizing a lower-bound on the {\it $\log$-full expected utility} using the same approach \citep{lacoste2011approximate, kusmierczyk2019variational}:
\begin{flalign}
\log \EEutil_p(a) &\ge \int q(\param)\cdot \log\left(\frac{p(\param\mid\data)\Util(a, \param)}{q(\param)}\right)\,d\param \\
\text{s.t.}\quad q_\text{LossVI}(\param) &= \underset{q}{\arg\min}\: D_{KL}\left(\tilde p_a(\param)\,\|\, q(\param)\right), \:\:\text{where}\:\: \tilde p_a(\param) = \frac{p(\param\mid\data)\Util(a, \param)}{\EEutil_p(a)} \label{eq:lossvi-kl}
\end{flalign}
The resulting bound is exactly the variational inference objective, but minimizes the KL divergence from a {\it utility-weighted} posterior $\tilde p_a(\param)$.

Implementation of LossVI requires joint optimization of the approximation $q(\param)$ and the actions $a(\cdot)$, and \citep{lacoste2011approximate} propose an EM procedure that alternates coordinate descents between the two, with applications to Gaussian processes.

\section{Loss-calibrated expectation propagation}

Expectation propagation (EP) approximates the posterior by decomposing the global inference into a collection of local inferences by message-passing \citep{minka2001expectation, minka2001ep, gelman2014expectation}. These local inferences operate on each term of some factorization of the posterior, typically the product of the prior and the likelihoods of each of the i.i.d. data:
\begin{flalign}
    p(\param\mid\mathcal{D}) = \frac{1}{Z}p_0(\param) \prod_i p(\data_i\mid\param) &\propto p_0(\param) \prod_i t_i(\param)\quad\approx\quad p_0(\param) \prod_i \tilde t_i(\param) = q(\varphi)
\end{flalign}
Each likelihood term $t_i(\param)$ is approximated in EP by a site potential $\tilde t_i(\param)$, such that the approximation shares the factorization of the true posterior. EP iteratively performs local approximate inferences for each site $\tilde t_i(\param)$, over three steps---Deletion, Projection, and Update---that replaces one site with its true likelihood term, optimizes a new approximation that matches moments to this hybrid distribution, and updates the original site accordingly ({\it cf.} box below, Algorithm 1). All of these operations are powered by properties of distributions in the same exponential family (EF), for which purpose we assign all sites to have the (unnormalized) exponential family form
\begin{flalign}
\tilde t_i(\param) \propto \exp\left(\langle \mathbf{s}(\param), \boldsymbol{\theta}\rangle\right) \label{eq:site-ef}
\end{flalign}
where $\mathbf{s}(\param)$ are the sufficient statistics and $\boldsymbol{\theta}$ are the natural parameters \citep{seeger2005expectation,nielsen2009statistical}. 

These updates are repeated across all site potentials until convergence. While these local inferences only minimize the forward KL divergence from $N$ different tilted distributions, this iterative message-passing procedure approximates the global minimization forward KL divergence from the posterior, $D_{KL}(p(\param\mid\data\,\|\,q(\param))$ \citep{minka2001ep}. Convergence is not guaranteed, but empirically EP performs well. 

Our contribution begins with the observation that the full expected utility in (\ref{eq:full-expected-util}) admits a factorization similar to that utilized by EP:
\begin{flalign}
\EEutil_p(a) = \int \frac{1}{Z} p_0(\param)\prod_n p(y_n\mid\bx_n,\param) \Util(\param, a) df
&\propto \int p(\param)\prod_n t_n(\param) t_\ell(\param) d\param
\end{flalign}
where the the utility function $\Util(a, \param)$ becomes the $N+1$\textsuperscript{th} term $t_\ell(\param)$.
As before, we approximate the $N$ likelihood terms and the new utility term with $N+1$ exponential family site potentials $\tilde t_i(\param)$ and $\tilde t_\ell(\param)$, respectively. The resulting loss-calibrated EP approximation $\overline{q}(\param)$ takes the form
\begin{flalign}
\overline{q}(\param)= p_0(\param)\prod_i \tilde{t}_i(\param) \tilde{t}_\ell(\param) \propto q(\param) \tilde{t}_\ell(\param)
\end{flalign}
and approximates the same utility-weighted posterior $\tilde p_a(\param)$ as LossVI in eq. \eqref{eq:lossvi-kl}. The approximation to the posterior $p(\param\mid\data)\approx q(\param)$ is a partial product of the site potentials that comprise $\overline{q}(\param)$: all except the utility site $\tilde t_\ell(\param)$. EP updates for the $N$ site approximations $\tilde t_i(\param)$ follow the standard EP procedure; the update for the new site approximation $\tilde t_\ell(\param)$ requires one extra computation. We present a single site update below in full detail, and give the aggregate LossEP procedure in Algorithm 1.

\begin{tcolorbox}[center, breakable, width=0.95\textwidth, colback={pale-gray}, boxrule=0pt] 
   {\bf Deletion:} Divide the utility site $\tilde t_\ell(\param)$ from $\overline{q}(\param)$ to form the cavity distribution, which {\it for the utility site} is identically the posterior approximation:
    \begin{flalign}
        q\scav{\ell}(\param) &= \overline{q}(\param)/\tilde t_\ell(\param) \propto p_0(\param) \prod_{i} \tilde t_i(\param) \propto q(\param)
    \end{flalign}
    In natural parameters, $\boldsymbol{\theta}\scav{\ell} = \overline{\boldsymbol{\theta}} - \boldsymbol{\theta}_\ell$.
    
   {\bf Projection:} Reinsert the true term $t_\ell(\param)$ into $q\scav{\ell}(\param)$ to form the tilted distribution, and project back into the approximating family:
    \begin{flalign}
        \overline{q}\new(\param) &= \proj\big[t_\ell(\param)q\scav{\ell}(\param)\big] \triangleq \underset{\overline{q}}{\arg\min}\:D_{KL}(t_\ell(\param)q\scav{\ell}(\param)\,\|\, \overline{q}(\param))
    \end{flalign}
    In natural parameters, this divergence minimization is equivalent to the gradient update $\overline{\boldsymbol{\eta}}\new = \boldsymbol{\eta}\scav{\ell} + \nabla_{\boldsymbol{\theta}\scav{i}}\log Z_\ell$, where $\boldsymbol{\eta}=\EE[\mathbf{s}(\param)]$ are the mean parameters and and
    \begin{flalign}
        \log Z_\ell = \log \int t_\ell(\param)q\scav{\ell}(\param)\,d\param = \EEutil_q(a)
    \end{flalign}
    is the log-normalizer of the tilted distribution. {\it For the utility site}, the true term $t_\ell(\param)=\Util(a, \param)$ is the full predictive utility and the log-normalizer is identically the $q$-full expected utility. To resolve the dependence on actions $a$, introduce the following prerequisite step:
    
    {\bf Action Selection:} [Utility site only] Select the $q\cav{\ell}$-optimal actions
   \begin{flalign}
   a_{q}&=\underset{a}{\arg\max}\:\EEutil_{q\cav{\ell}}(a)
   \end{flalign}
   Performing this step only requires information already available at the local site update.
    
    {\bf Update:} Divide the utility-cavity distribution from the projected $\overline{q}\new(\varphi)$ to back out the updated site $\tilde t_\ell(\varphi)$. With damping $\delta\in[0, 1)$,
    \begin{flalign}
        \tilde{t}_\ell\new(\param) &= (\overline{q}\new(\param)\,/\, q\scav{\ell}(\param))^\delta\cdot \tilde{t}_\ell(\param)^{1-\delta}
    \end{flalign}
    In natural parameters, $\boldsymbol{\theta}_\ell\new = \delta(\overline{\boldsymbol{\theta}}\new - \boldsymbol{\theta}\scav{\ell})+(1-\delta)\boldsymbol{\theta}_\ell$.
\end{tcolorbox}

Only the tilted distribution of the utility site is an explicit function of actions $a$, presenting the only opportunity to optimize actions $a$ during LossEP. Performing action selection at this site maximizes the $q$-gain, which is consistent with other loss-calibrated methods, albeit motivated more naturally in this context. Critically though, loss-calibration is still occuring at {\it all} site updates, as the Gaussian approximation to the utility $\tilde t_\ell(\param)$ is present in every cavity distribution $q\scav{i}(\param)$, and it contains information about the decision-making process. 

{\SetAlgoNoLine \DontPrintSemicolon
\begin{algorithm}
 \textit{Given:} Dataset of $N$ i.i.d. datapoints $\data=\{\bx_i,y_i\}$, Model with posterior over parameters $p(\param\mid\data)$ that admits factorization of the form $p(\param\mid\data)\propto p_0(\param) \prod_i t_i(\param)$, Decision problem with utility function $\util(a(\cdot),\cdot)$\;
 Initialize site approximations $\tilde t_i(\param)$ and $\tilde t_\ell(\param)$ {\it s.t.} $\overline{q}(\param)=p_0(\param)\prod_{i=1}^N \tilde t_i(\param)\tilde t_\ell(\param)$ approximates the utility-weighted posterior $p(\param\mid\data)\Util(a,\param)$\;
 \Repeat{convergence}{
    {\it In random order:}\;
 	For each $n$\;
	(\textbf{Deletion}) Compute the cavity distribution $q\scav{i}(\param)=\overline{q}(\param)/\tilde t_i(\param)$\;
	(\textbf{Projection}) Minimize $\overline{q}^{\text{new}}(\param)=\proj[t_i(\varphi)q\scav{i}(\param)]$ by matching moments\;
	(\textbf{Update}) Back out the site update $\tilde t_i^\text{new}(\param)=(q^{\text{new}}(\param)/q\scav{i}(\param))^\delta\cdot \tilde t_i(\param)^{1-\delta}$\;
	For $\ell$\;
	(\textbf{Deletion}) Compute the cavity distribution $q\cav{\ell}(\param)=\overline{q}(\param)/t_\ell(\param)$\;
	(\textbf{\textit{Action Selection}}) Select the $q\scav{\ell}$-optimal actions $a_{q}={\arg\max}_a\:\EEutil_{q\cav{\ell}}(a)$\;	
	(\textbf{Projection}) Minimize $\overline{q}^{\text{new}}(\param)=\proj[U(a_q, \param)q\scav{\ell}(\param)]$ by matching moments\;
	(\textbf{Update}) Back out the site update $\tilde t_\ell^\text{new}(\param)=(\overline{q}^{\text{new}}(\param)/q\scav{\ell}(\param))^\delta\cdot \tilde t_\ell(\param)^{1-\delta}$\;
 }
 \caption{Loss-calibrated expectation propagation}
\end{algorithm}
}

\subsection{Example: the clutter/nuclear reactor problem}

To illustrate loss-calibration, we repurpose the ``nuclear reactor'' problem used by \citep{lacoste2011approximate} and \citep{abbasnejad2015loss} in those works on loss calibration. Conveniently, this decision problem can be superimposed on the generative model of the clutter problem used in \citep{minka2001expectation} in that work on expectation propagation. We consider a decision-maker receiving noisy measurements of the temperature $\param$ of a nuclear reactor, tasked with shutting it down ($a=1$) if that temperature has exceeded a ``meltdown'' point ($\varphi>T_\text{crit}$) or leaving it on ($a=0$) otherwise. Defining $\util_{ij}$ as the utility received for taking action $a=i$ when the state $\II[\varphi>T_\text{crit}]=j$ was true \citep{poor2013introduction}, this decision problem features a natural asymmetry, with a maximal penalty for the False Negative case of leaving it on during a meltdown ($\util_{01}= 0$) and a modest penalty for the False Positive of shutting down despite normal operation ($\util_{10}$). 

After a small number of observations, suppose we have a complicated, bimodal posterior distribution $p(\param\mid\data)$ with the minor mode informing a potential meltdown (Figure 1). Under a Gaussian approximation, we may underrepresent or ignore this probability mass, and select the suboptimal action of ``leaving it on'' when the Bayes-optimal action is to ``shut it down'' (see figure legend). For not only VI and EP {\it but also} LossVI, this is exactly what happens (Figure 1; blue curves). Only LossEP captures the tail and can select the Bayes-optimal action (Figure 1; red curves). It is a poor visual approximation of the true posterior, but when our goal is decision-making, this is not important.

\begin{figure}[b]
\centering
\includegraphics[width=0.45\textwidth]{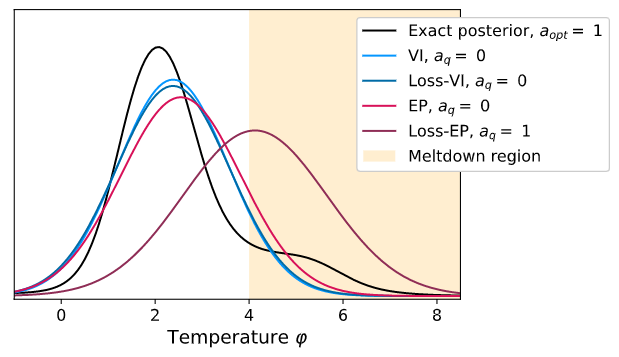}
\caption{{\bf Nuclear reactor decision problem for a small dataset observed under the clutter problem.} A decision-maker must act to leave the reactor running or shut it down, based on whether the true mean temperature, inferred from that dataset, is greater than a critical meltdown threshold. Given this bimodal posterior distribution (black), the optimal action shuts it down (see legend, $a_\text{opt}=1$). However, the Gaussian approximations from VI, LossVI, and EP (pale blue, dark blue, and pale red, {\it resp.}) do not capture the second mode and do not select this action ($a_q=0$). The Gaussian approximation from LossEP (dark blue) skews to overrepresent the second mode, and selects the optimal action ($a_q=1$).}
\label{fig:clutter}
\end{figure}


\section{Applications to data with Gaussian process models}

We explore the differences in the approximations achieved between loss-calibrated EP and standard EP, as well as VI and LossVI, using a simulated decision-making problem with asymmetric utility functions. When the inference and/or decision-making is difficult enough to yield loss of information and/or ambiguous predictions, we suspect that our loss-calibrated methods will be most differentially successful. We use Gaussian process classification in all cases for its analytical tractability. We present two applications: the first is a minimal two-point dataset \citep{nickisch2008approximations} to visualize the various approximations, and the second a sweep of different utility functions and predictive datasets \citep{lacoste2011approximate}. Together, they show that loss-calibration---at least within this context---can yield modest increases in performance, as measured by utility.

\StopCensoring 
We implement all of EP, and LossEP in python using JAX \citep{jax2018github}, and MCMC, VI, and LossVI in numpyro \citep{phan2019composable, bingham2019pyro}. 
To begin, we derive the relevant equations for loss-calibrated inference in the GP context, with notation following \citet{rasmussen2005gaussian}.

\subsection{Approximate inference and loss-calibration for Gaussian process classifiers}

Approximate inference in GP classification provides an approximation to the intractable posterior $p(\bff\mid X, \by)\xrightarrow{\text{approx.}}q(\bff\mid X,\by)=\Nrm(\bff;\, \bmu_q, \Sigma_q)$, where the latent function at the observed datapoints $\bff$ assumes the role of the parameters $\param$. We assume the probit likelihood model throughout such that all integrals can be resolved in closed-form. The prior, likelihood, and posterior take the form
\begin{flalign}
\nonumber p(\bff\mid X) &= \Nrm(\bff;\, 0, \Kxx),\quad
p(\by\mid\bff) = \prod_{i=1}^N \Phi(y_i f_i),\quad
p(\bff\mid X, \by) = \frac{p(\by\mid\bff)p(\bff\mid X)}{p(\by\mid X)}
\end{flalign}
where $K$ is the covariance kernel which, in our case, is the radial basis function kernel defined by 
\begin{flalign}
K_{ij}=k_\xi(\bx_i, \bx_j)=\sigma^2\exp(- \|\bx_i-\bx_j\|^2_2 / 2\ell^2)
\end{flalign} 
with variance and lengthscale hyperparameters $\xi=\{\sigma^2,\ell\}$. To make predictions for latent function values outside the dataset, we can marginalize the product of any approximation $q(\bff)$ and the conditional distribution, yielding the posterior predictive distribution.
\begin{flalign}
p_q(\fs\mid X, \by, \bxs) &= \int d\bff\, p(\fs\mid X, \bxs, \bff) q(\bff\mid X, \by) = \Nrm(\fs;\, m_{\fs}, v_{\fs})
\end{flalign}
\vspace*{-18pt}
\begin{flalign}
\text{where}\quad m_{\fs} &= \ksx\Kxx\inv\bmu_q \label{eq:postpred-mean}\\
v_{\fs} &= \kss-\ksx\Kxx\inv\kxs+\ksx\Kxx\inv\Sigma_q\Kxx\inv\kxs \label{eq:postpred-var}
\end{flalign}
and $\ksx$ is the predictive covariance kernel, with ${\ksx}_j = k_\xi(\bxs, \bx_j)$. 

The expected utility depends on the marginal predictive distribution, which is available in closed-form for the probit likelihood. As above, we marginalize the product of the probit likelihood and the posterior predictive distribution, yielding
\begin{flalign}
p_q(\ys\mid X, \by, \bxs) &= \int d\fs\, p(\ys\mid\fs) p_q(\fs\mid X, \by, \bxs) = \Phi\left(\ys\cdot \frac{m_{\fs}}{\sqrt{1+v_{\fs}}} \right)  \label{eq:gpc-post-pred}
\end{flalign}

For action selection in GPC, we can rearrange the order of integration in  \eqref{eq:gpc-post-pred} of the conditional expected utility in \eqref{eq:cond-expected-util}, reducing it to a single integral of this predictive likelihood and the utility function \citep{vehtari2012survey}, which decomposes into four pieces for each of the four outcomes. We define them as $u_{00}$, $u_{01}$, $u_{10}$, and $u_{11}$, where $u_{ij}$ is the utility of selection action $i$ when outcome $j$ occurred \citep{poor2013introduction}.
\begin{flalign}
\EEutil_q(a(\bxs)\mid\bxs) &= \int d\ys\,p_q(\ys\mid\bxs,X, \by)\,\util(\ys,a(\bxs)) \label{ep:gpc-conditional-util}
\\ 
&= \II[a=+1]\left(u_{10}\Phi(-z)+u_{11}\Phi(z)\right) + \II[a=-1]\left(u_{00}\Phi(-z)+u_{01}\Phi(z)\right) \\
\nonumber &\quad\text{where}\quad z=\frac{m_{\fs}}{\sqrt{1+v_{\fs}}}
\end{flalign}
and $\II[\cdot]$ is an indicator function with value $1$ if the condition is true and $0$ otherwise. The conditional expected utility, when minimized pointwise, minimizes the full expected utility \citep{berger2013statistical}. No matter the procedure for selecting $q(\bff)$, we select actions by minimizing this term, which generates the following $q$-action in closed form:
\begin{flalign}
a_q(\bxs) &= \text{sign}\left(\frac{m_{\fs}}{\sqrt{1+v_{\fs}}} - b\right),\quad b=\Phi\inv\left(\frac{u_{00} - u_{10}}{(u_{00}- u_{10}) - (u_{01}- u_{11})}\right)
\end{flalign}
The "bias" $b$ reflects the asymmetry of the utilities, and acts to move the decision boundary / threshold up if its more favorable to select $-1$ or down if its more favorable to select $+1$.

\subsection{Visualizing a two-point dataset}

Suppose we had a dataset of only two points, $\data=\{(-\sqrt{2},-1), (\sqrt{2},+1)\}$. In this case, we can compute the exact posterior $p(\bff\mid X,\by)$ on a grid and plot it against the various approximations. We select hyperparameters $\log \sigma=1.5$ and $\log \ell=1.0$ (cf. \citep{nickisch2008approximations}, Figs. 3, 4) and a decision problem with an asymmetric utility function: $u_{00}=u_{11}=1.0$, $u_{01}=0.0$, and $u_{10}=0.5$ (cf. \citep{lacoste2011approximate}, Table 4). 

\begin{figure}
    \centering
    \includegraphics[width=0.8\textwidth]{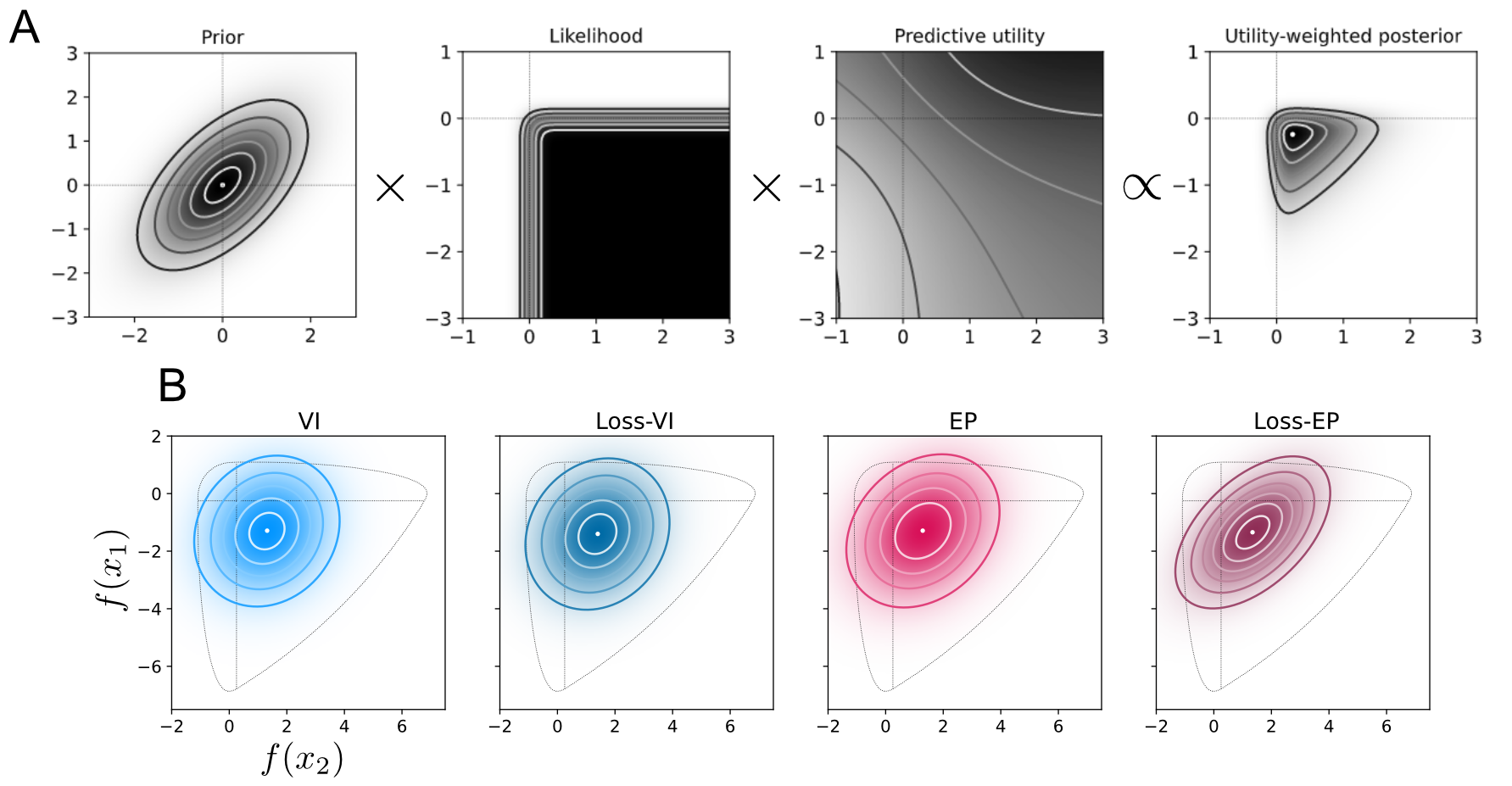}
    \caption{{\bf Loss-calibrated approximate inference for GP classification on a two-point dataset}.  {\bf A}, The utility-weighted posterior approximated by loss-calibration, achieved by multiplying the jointly Gaussian GP prior and the probit likelihood for the two datapoints, $(-\sqrt{2}, -1)$ and $(\sqrt{2},+1)$, and the predictive utility. {\bf B}, Approximations to the posterior for VI (pale blue) EP (pale red), and to the utility-weighted posterior for loss-calibrated VI (dark blue) and loss-calibrated EP (dark red).}
    \label{fig:nr-2point}
\end{figure}

Together, these induce a non-Gaussian posterior---the product of a joint Gaussian prior with steep, sigmoidal likelihoods yields a posterior with tails strongly skewed into one quadrant. The predictive utility skews this posterior very slightly towards the upper right quadrant, as overestimating the latent function's values is more favorable under the utility function which disproportionately penalizes underestimating (Fig. \ref{fig:nr-2point}A). Compared to standard EP and VI, their loss-calibrated counterparts do not modify the means, and LossEP increases the covariance estimate (Fig. \ref{fig:nr-2point}C).

\subsection{Assessing the impact of loss-calibration under asymmetric utilities}

We continued with a quantitive performance test of Loss-EP versus EP as well as Loss-VI and VI (Table \ref{tab:util-asymm}). We simulated datasets as $N=15$ datapoints $\bx_i$ sampled uniformly from $[-10, 10]$, followed by draws from the GP prior and binary observations $y_i$ under the GPC model with hyperparameters as above. Action selection was performed for each of five increasingly asymmetric utility functions, with the False Negative utility always taking the minimum-possible value $u_{01}=0.0$ and the False Positive utility ranging from minimum-possible $u_{10}=0.0$ to nearly maximum-possible $u_{10}=0.95$. Utilities were evaluated at 1000 predictive datapoints uniformly sampled from each of $[-10, 10]$, $[-8, 12]$, or $[-5, 15]$, increasing covariate shift from the observed datasets. Using this finite dataset for prediction can be viewed in the transductive scenario in which we are interested in evaluating the utility {\it at these points}. This is beneficial, as it reduces the unknown data-generating distribution $p(\bxs)$ to a trivial ``comb'' distribution over that dataset $p(\bxs)=\sum_{c} \delta(\bxs - {\bxs}_c)$. The resulting experiment is identical to that used in \citep{lacoste2011approximate}.

The performance is the full expected utility $\EEutil_p(a_q)$ of the actions selected under any of the four approximations $q$: EP, Loss-EP, VI, and Loss-VI. This utility is at best $\EEutil_p(a_\text{opt})$---the utility of the Bayes-optimal action---and is at worst $\EEutil_p(-a_\text{opt})$---the utility of always taking the opposite of the Bayes-optimal action. The following metric normalizes the utility discrepancy in eq. \eqref{eq:regret} from $0$ to $1$:
\begin{flalign}
\widetilde U(q) = \frac{\EEutil_p(a_\text{opt}) - \EEutil_p(a_q)}{\EEutil_p(a_\text{opt}) - \EEutil_p(-a_\text{opt})} \label{eq:utility-metric}
\end{flalign}
Different datasets vary in their best-case and worst-case expected utilities, and the metric permits meaningful averaging across datasets. We report this utility metric for each inference method, utility function, and covariate shift as averages over 20 simulated datasets each in Table \ref{tab:util-asymm}.

Recall that the motivation of loss-calibrated approximate inference was that the expected utility $\EEutil_p(a_\text{opt})$ is difficult to compute. For GPC, we can compute it using MCMC \citep{hoffman2014no}, generating samples $\bff\sim p(\bff\mid X, \by)$ and computing an sample-based estimate of the conditional expected utility in \eqref{eq:full-expected-util}. Averaging over the dataset defining $p(\bxs)$, we recover an estimate of $\EEutil_p(a)$.
We can either optimize this function for $a$ to estimate $a_\text{opt}$ (and $-a_\text{opt}$) or plug in $a_q$ for one of the approximations $q$ to estimate $\widetilde{U}(q)$ for each of VI, LossVI, EP, and LossEP.

All methods report near-ceiling performance across all utility functions and degrees of predictive covariate shift, though EP and LossEP consistently outperform VI and LossVI by several orders of magnitude (Table \ref{tab:util-asymm}). While LossVI does not outperform VI in any case, LossEP offers very small improvements over EP when the utility asymmetry is moderate and covariate shift is {\it not} large, echoing the results visualized in the previous section {\it and} those observed by \cite{lacoste2011approximate} in their original report. No differences between EP and Loss-EP or VI and Loss-VI were significant by a Wilcoxon signed-rank test after corrections for multiple comparisons. 

\begin{table}
\centering
\includegraphics[width=0.85\textwidth]{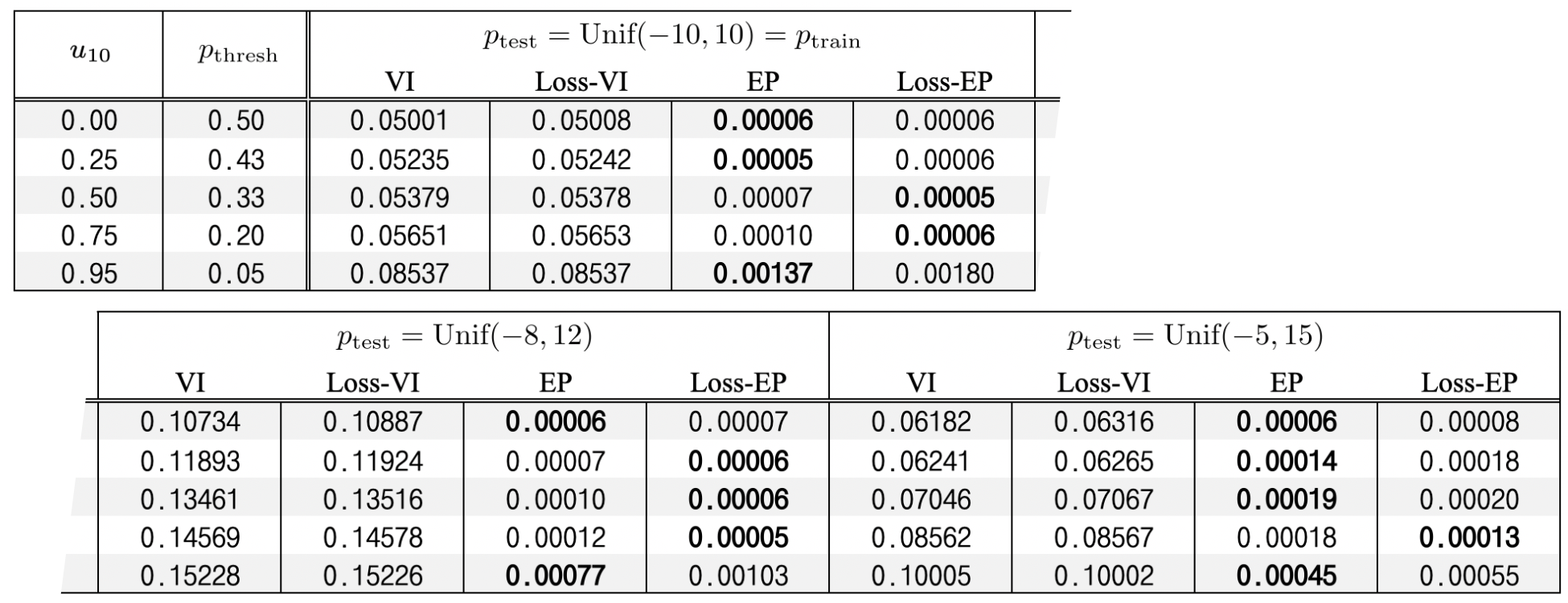}
\caption{{\bf Performance comparison of loss-calibrated methods on simulated datasets.} Expected utility from actions selected under approximate posterior distributions from each of VI, loss-calibrated VI, EP, and loss-calibrated EP, for utility functions of increasing asymmetry and predictive datasets with increasing covariate shift. Performance is reported using the normalized expected utility metric in eq. \eqref{eq:utility-metric}, averaged over 20 repeats of the simulation. Zero is Bayes-optimal performance, and does not mean perfect performance. EP is consistently better than both VI-type methods, and LossEP is better still for moderate utility asymmetries. Bolded results are best performances within each experimental category.}
\label{tab:util-asymm}
\end{table}



\section{Discussion and future work}

With this paper, we introduced loss-calibrated expectation propagation, diversifying the suite of loss-calibrated inference methods to include the message-passing-like iterative computations of EP. We showed that the factorization of the posterior distribution approximated by EP can be repurposed to a factorization of the expected utility, and appends one site factor to the resultant approximation: $\tilde t_\ell(\param)$ which approximates the predictive utility function $\Util(a,\param)$. When iteratively updating all data site factors $\tilde t_i(\param)$, the utility site factor $\tilde t_\ell(\param)$ is included in all cavity distributions $q\scav{i}(\param)$, making all sites ``aware'' of the utility function. When iteratively updating the utility site itself, the tilted distribution's log-normalizer is precisely the $q$-expected utility, providing a principled justification of optimizing $\EEutil_q(a)$ as a surrogate for the true expected utility $\EEutil_p(a)$. As a bonus, action selection can be transparently performed within this site update. 
In these ways, Loss-EP shares the elegant, intuitive, and desirable theoretical foundations of loss-calibrated approximate inference more broadly. 

That said, it also shares the situational, modest performance increases in practice observed in previous works, which spans not only Gaussian process models \citep{lacoste2011approximate} but also Bayesian neural networks \citep{cobb2018loss} and continuous-valued estimation tasks \citep{kusmierczyk2019variational} not considered here. Using {\it any} EP-type approximation, loss-calibrated or not, does grant dramatic performance improvements, however, and is an important, novel observation from our efforts. 
Curiously, we observe that there is likely more than one way to develop a loss-calibrated inference procedure around the flexible machinery of expectation propagation, and future work could better harness the desirable computational properties commonly associated with EP to improve performance.

Following \citep{kusmierczyk2020correcting}, future work could develop an EP-type approximation to Generalized Bayesian Inference \citep{bissiri2016general}, which updates prior beliefs about a parameter of interest that is connected to our observations by arbitrary loss functions, rather than likelihood functions. Following the substantial body of theory work into loss functions and machine learning \citep{reid2011information}, we could perform iterative EP updates that minimize not the KL divergence, but instead some general divergence measure (also suggested in \citep{gelman2014expectation}) that is uniquely determined by the loss function in the classification case. This could be considered optimization in the natural ``geometry'' of the decision problem \citep{dawid2005geometry}, of which a certain subclass of problems are well-defined \citep{dawid1994proper}.

\subsection{Related work}

Loss-calibrated approximate inference is only one family of approaches for ``loss-aware'' model learning. Direct loss minimization \citep{wei2021direct} directly minimizes expected loss, but in a constrained optimization akin to rate-distortion theory \citep{cover1999elements}, which has a natural connection to variational inference \citep{alemi2018fixing}. Accuracy-versus-uncertainty calibration \citep{krishnan2020improving} provides a heuristic objective for training deep networks to enforce better-calibrated estimators. Target-aware Bayesian inference \citep{rainforth2020target} formalizes an sampling-based approach to estimating the expected utility with importance sampling (IS). Loss-calibrated Monte Carlo action selection \citep{abbasnejad2015loss} also optimizes an importance sampling proposal distribution, but minimizing the probability of selecting suboptimal actions with minimal number of samples. 

This last work echoes the deep theory of surrogate loss functions from statistical learning theory, which minimize a bound on the generalization error (equivalently, the 0-1 loss, or the probability of error) that depends directly on the utility/loss function of the decision problem \citep{bartlett2006convexity, steinwart2008support, reid2011information}. \citet{nguyen2009surrogate} even select actions according to an approximation, and such results hinge on an expected loss {\it over datasets}. In contrast, our complementary Bayesian viewpoint has emphasized how to make optimal decisions conditionally on a given dataset rather than in expectation over all possible datasets.

\section*{Broader Impact}

Optimal decision-making under specific design constraints (as represented by a utility function and/or tolerances on error rates) is a fundamental aspect of fair and safe applications of ML to real-world data. While loss-calibration as we present it here is a generic theoretical tool, principles such as these could be leveraged to better-constrain the decision-making faculties of modern Bayesian ML tools. For example, it could be productive for researchers to think explicitly about the costs/utilities of decision-making in their applications {\it at inference time}. Because Bayesian decision theory only prescribes optimal decision-making under exact Bayesian inference, a modernized theory will be critical as approximate methods for complex models become more and more prevalent.



\begin{ack}

MJM was supported by an NSF Graduate Research Fellowship. 


\end{ack}

\bibliography{losscal}
\bibliographystyle{unsrtnat}

\newpage
\appendix

\section*{{\it Supplemental Materials for:} Loss-calibrated expectation propagation for approximate Bayesian decision-making}



\section{Analytical EP and Loss-EP updates for the clutter/reactor problem}

Here, we describe EP and Loss-EP for the clutter and reactor examples that we use in Fig. \ref{fig:clutter}.

The "clutter problem" was the guiding example used by Minka in the original paper introducing expectation propagation, and consequently is often the gateway example for hopeful users of EP. In practice, the clutter problem is a two-component Bayesian mixture-of-Gaussians, with all parameters fixed in advance except for the latent mean of one of the components. In this way, inference in the clutter problem can be conceptualized as both a signal detection and/or latent variable model, in which we learn from noisy observations the mean of a signal buried amongst "clutter" (noise).

We define the observations as $y$, the signal mean as $\param$, the proportion of clutter as $\pi$, and the clutter variance as $v_c$. The resulting likelihood function for $y\mid\param$, alongside a Gaussian prior over $\param$, defines the clutter problem's generative model:
\begin{flalign}
    p_0(\param) &= \Nrm(\param;\, 0, v_0) \\
    p(y\mid\param) &= (1-\pi)\Nrm(y;\, \param, 1) + \pi\Nrm(y;\, 0, v_c) \label{eq:exact-clutter-lik}
\end{flalign}
where $\Nrm(\param; a, A)$ defines a Gaussian distribution over $\param$ with mean $a$ and variance $A$. We use the same values used by Minka: $\pi=0.5$, $v_c=10$, and $v_0=100$ (a non-informative prior); however, we assume scalar observations to simplify algebra and visualization.

The clutter problem is deceivingly simple to write down. However, given a dataset of observations $\data=\{y_i\}_{i=1}^N$, the posterior is a mixture of $2^N$ Gaussians\footnote{To see why, note that the total data likelihood is $\prod_i\left((1-\pi)\Nrm(y_i;\, \param, 1) + \pi\Nrm(y_i;\, 0, v_c)\right)$ which, if we multiply out each binomial, yields $2^N$ terms of the form $c\cdot\Nrm(y\; ...)\Nrm(y\; ...)$.}, and it is exponentially difficult to compute exact moments e.g., if attempting belief propagation, as was the motivation for EP at the time. This posterior becomes the target of our EP approximation: 
\begin{flalign}
    p(\param\mid\mathcal{D}) = \frac{1}{Z}p_0(\param) \prod_i p(\data_i\mid\param) &\propto p_0(\param) \prod_i t_i(\param) \\
    &\approx p_0(\param) \prod_i \tilde{t}_i(\param) = q(\param)
\end{flalign}
where $t_i(\param)$ is the exact likelihood factor in \eqref{eq:exact-clutter-lik}, $\tilde{t}_i(\param)=\Nrm(\param;\, m_i, v_i)$ is its Gaussian approximate factor in the EP approximation. The prior $p_0(\param)$ is Gaussian and needs no approximation.

EP in the clutter problem can be performed analytically. We present the Deletion-Projection-Update steps using the mean and variance $(m, v)$ on the left, and using the natural and mean parameters $(\nat, \Nat)$ and $(\mean, \Mean)$, resp., on the right to emphasize the exponential family perspective. Note that only \eqref{eq:clutter-tilted} and its gradients are specific to the clutter problem; the rest is fully general.

{\bf Deletion}:
\begin{equation}
\begin{aligned}[t]
q\scav{i}(\param) &= q(\param)\,/\, \tilde{t}_i(\param) \\
v\scav{i} &= \left(v\inv - v_i\inv\right)\inv \\
m\scav{i} &= v\scav{i}\left(v\inv m - v_i\inv m_i\right)
\end{aligned}
\qquad 
\begin{aligned}[t]
q\scav{i}(\param) &= q(\param)\,/\, \tilde{t}_i(\param) \\
\Nat\scav{i} &= \Nat - \Nat_i \\
\nat\scav{i} &= \nat - \nat_i
\end{aligned} \label{eq:clutter-deletion}
\end{equation}

{\bf Projection}:
The projection step hinges on the log-normalizer of the tilted distribution, which for the clutter problem is given by
\begin{flalign}
Z_i = \int d\param\, q\scav{i}(\param)p(\data_i\mid\param) &= \int d\param\, \Nrm(\param;\, m\scav{i}, v\scav{i})\left((1-\pi)\Nrm(y;\, \param, 1) + \pi\Nrm(y;\, 0, v_c) \right) \\
&= (1-\pi)\Nrm(y;\, m\scav{i}, 1+v\scav{i}) + \pi\Nrm(y;\, 0, v_c) \label{eq:clutter-tilted}
\end{flalign}
where the first term follows from the following Gaussian refactoring of prior-times-likelihood into evidence-times-posterior. We can then integrate out the posterior:

{
    \color{sidenote-gray}
    {\it --- Multiplication of two Gaussians (prior $\times$ likelihood = evidence $\times$ posterior)}
    \begin{flalign}
        \Nrm(x;\, a, A)\Nrm(x;\, b, B) &= \Nrm(x;\, a, A)\Nrm(b;\, x, B)\quad\text{(symmetry)}\nonumber \\
        &= \Nrm(a;\,b, A+B)\Nrm(x;\, B(A+B)\inv a+A(A+B)\inv b, A(A+B)\inv B) \label{eq:normal-times-normal}
    \end{flalign}
}

\begin{equation}
\begin{aligned}[t]
q\new(\param) &= \proj\big[q\scav{i}(\param)p(\data_i\mid\param)\big] \\
m\new &= m\scav{i} + v\scav{i}\nabla_{m\scav{i}}\log Z_i \\
v\new &= v\scav{i} + v\scav{i}\left(2\nabla_{v\scav{i}}\log Z_i - (\nabla_{m\scav{i}}\log Z_i)^2\right)v\scav{i}
\end{aligned}
\qquad 
\begin{aligned}[t]
q\new(\param) &= \proj\big[q\scav{i}(\param)p(\data_i\mid\param)\big] \\
\mean\new &= \mean\scav{i} + \nabla_{\nat\scav{i}}\log Z_i \\
\Mean\new &= \Mean\scav{i} + \nabla_{\Nat\scav{i}}\log Z_i
\end{aligned} \label{eq:clutter-projection}
\end{equation}
The projection operator $\proj$ is a common shorthand for KL divergence minimization, i.e. $\proj[p(\param)]=\arg\min_q D_{KL}(p(\param)\,\|\,q(\param))$, which for an exponential family $q$ is equivalent to moment-matching to $p$. These moments, which we represent as the mean parameters of $q$, take the form above due to specific exponential family properties similar to exponential tilting \citep[for more details, cf.][]{seeger2005expectation}. We can manually compute the gradients of the $log$ of \eqref{eq:clutter-tilted} to yield analytical updates
\begin{flalign}
m\new &= m\scav{i} + v\scav{i}\cdot r_i \frac{y_i-m\scav{i}}{1+v\scav{i}} \\
v\new &= v\scav{i} - {v\scav{i}}^2\cdot \left(r_i \frac{1}{1+v\scav{i}} - r_i(1-r_i) \frac{(y_i-m\scav{i})^2}{(1+v\scav{i})^2}\right) \\
&\text{and}\quad r_i=\frac{1-\pi}{Z_i}\Nrm(y_i;\,m\scav{i}, 1+v\scav{i})
\end{flalign}
In practice, we found it easier and more concise to compute automatic gradients with modern ML tools such as \texttt{JAX} directly on \eqref{eq:clutter-tilted}, but we provide these analytical solutions in terms of mean and variance for completeness. Because these are very short expressions, we observed that the automatic gradients were accurate up to numerical precision.

{\bf Update}:
\begin{equation}
\begin{aligned}[t]
\tilde{t}_i\new(\param) &= (q\new(\param)\,/\, q\scav{i}(\param))^\delta\cdot \tilde{t}_i(\param)^{1-\delta} \\
v_i\new &= (\delta(v^{\text{new},-1}- v^{\backslash i,-1})+(1-\delta)v_i\inv)\inv \\
m_i\new &= v_i\new(\delta(v^{\text{new},-1}m\new- v^{\backslash i,-1}m\scav{i}) \\
&\quad- (1-\delta)v_i\inv m_i)
\end{aligned}
\qquad 
\begin{aligned}[t]
\tilde{t}_i\new(\param) &= (q\new(\param)\,/\, q\scav{i}(\param))^\delta\cdot \tilde{t}_i(\param)^{1-\delta} \\
\Nat_i\new &= \delta(\Nat\new - \Nat\scav{i})+(1-\delta)\Nat_i \\
\nat_i\new &= \delta(\nat\new - \nat\scav{i})+(1-\delta)\nat_i
\end{aligned} \label{eq:clutter-update}
\end{equation}

The "reactor problem" is the guiding example used by Lacoste-Julien et al. in the original paper introducing loss-calibrated approximate inference, and conveniently sits atop the clutter problem with no modifications necessary. The reactor problem assigns a binary utility/cost function to whether a signal---putatively the internal temperature of a nuclear reactor---is above or below a critical "meltdown" temperature, and whether a user acts to keep the system on or turn it off. This real-world framing motivates use of a highly asymmetric utility/cost, in which it is catastrophically costly to keep the system on when the internal temperature is critical.

Atop the clutter problem, this temperature can be the latent mean itself $\param$. We define the utility of a true negative $\Util(\param<\Tcrit, a=\text{keep on})=L_0$, a false positive $\Util(\param<\Tcrit, a=\text{turn off})=L_1,$ a true positive $\Util(\param\ge\Tcrit, a=\text{keep on})=H_1$, and a false negative $\Util(\param\ge\Tcrit, a=\text{turn off})=H_0$. Following Abbasnejad et al., we choose $H_0\ll L_1 \le L_0 < H_1$, with the notational convenience $H_a$ to equal $H_1$ if $a=1$ and $H_0$ if $a=0$. 

Note that the clutter problem would defines $\Util(a, \param)$ directly with a single action $a$ over the entire dataset. We do not have a utility function of the form $\util(a(\cdot), \ys)$ with an action {\it function} as used in the main text. In essence, we infer: does this dataset $\data$ reflect a single latent signal whose mean "temperature" $\param$ is above the critical threshold $\Tcrit$? Clearly, this is a function of the posterior $p(\param\mid\data)$ that we are approximating with EP.

For loss-calibrated EP in the reactor problem, the only major change from the clutter problem is the projection step at the utility site (cf. Section 3, main text). This extension can substantially complicate computation in more elaborate decision problems, but reduces to a sum of Gaussian CDFs in our case.

{\bf Projection with loss-calibration}:
The projection step still hinges on the log-normalizer of the utility-tilted distribution, which for the reactor problem is given by
\begin{flalign}
Z_\ell &= \int d\param\, q\scav{i}(\param)U(a, \param) \\
&= L_a\cdot\int_{-\infty}^{\Tcrit} d\param\, \Nrm(\param;\, m\scav{\ell}, v\scav{\ell}) + H_a\cdot\int_{\Tcrit}^{\infty} d\param\, \Nrm(\param;\, m\scav{\ell}, v\scav{\ell}) \\
&= L_a\cdot\Phi(\Tcrit;\,m\scav{\ell},v\scav{\ell}) + H_a\cdot(1-\Phi(\Tcrit;\,m\scav{\ell},v\scav{\ell}))
\end{flalign}
The shorthand $\Phi(\tau;\, c, C)=\int_{-\infty}^\tau dy\,\Nrm(y;\, c,C)$ is the normal cdf with mean $c$ and [co]variance $C$, evaluated at $\tau$. With an analytic form for the utility-tilted distribution, we can compute automatic gradients as above and use the same projection equations in \eqref{eq:clutter-projection}. Deletion and update computations are identical to those in \eqref{eq:clutter-deletion} and \eqref{eq:clutter-update}, respectively.







\end{document}